\documentclass[10pt,twocolumn,letterpaper]{article}
\usepackage{pdfpages}
\usepackage{blindtext}
\usepackage{textcomp}
\usepackage{gensymb}
\usepackage{stackengine}
\usepackage{graphicx}
\usepackage{capt-of}
\usepackage{varwidth}
\usepackage{mathtools}
\usepackage{floatrow}
\usepackage{float}
\usepackage{esvect}
\usepackage{nicematrix}

\newfloatcommand{capbtabbox}{table}[][\FBwidth]
\usepackage{array}
\usepackage{nccmath}
\usepackage[export]{adjustbox}
\usepackage{booktabs,rotating,bigstrut}
\usepackage{bbm}
\usepackage{bm}
\newsavebox\tmpbox
\usepackage{scalerel,amssymb}
\usepackage{subcaption}
\usepackage{cuted}
\usepackage[font=small,labelfont=bf]{caption}
\usepackage{afterpage}
\usepackage{everypage}
\usepackage{arydshln}
\usepackage{multirow}

\newcommand{\RNum}[1]{\uppercase\expandafter{\romannumeral #1\relax}}

\def\msquare{\mathord{\scalerel*{\Box}{gX}}}
\DeclarePairedDelimiterX{\norm}[1]{\lVert}{\rVert}{#1}
\DeclareMathOperator*{\argmin}{\arg\!\min}
\DeclareMathOperator*{\argmax}{\arg\!\max}
\DeclareSymbolFont{symbolsC}{U}{pxsyc}{m}{n}
\DeclareMathSymbol{\coloneqq}{\mathrel}{symbolsC}{"42}

\usepackage{makecell}

\DeclareCaptionLabelFormat{andtable}{#1~#2  \&  \tablename~\thetable}

\newcommand{\captionaboveof}[3][]{%
    \vskip-\abovecaptionskip
    \vskip+\belowcaptionskip
    \def\@captype{#2}%
    \ifx\@nnil#1\@nnil
        \caption{#3}%
    \else
        \caption[#1]{#3}%
    \fi
    \vskip+\abovecaptionskip
    \vskip-\belowcaptionskip
}
\makeatother

\restylefloat{table}

\makeatletter
\newif\ifdelaymatch

\newcommand{\delayfloat}[4]{
  \ifnum\value{page}<#2\relax
    \afterpage{\delayfloat{#1}{#2}{#3}{#4}}%
  \else
    \delaymatchfalse
    \ifcase#3\relax\or
      \if@firstcolumn \delaymatchtrue \fi
    \or
      \if@firstcolumn\else \delaymatchtrue \fi
    \fi
    \ifdelaymatch
      \begin{#1}[t]
        \box#4
      \end{#1}
    \else
      \afterpage{\delayfloat{#1}{#2}{#3}{#4}}%
    \fi
  \fi}

\newif\ifdelaymatchtop

\newcommand{\delaytop}[2]{
  \ifnum\value{page}<#1\relax
    \afterpage{\delaytop{#1}{#2}}%
  \else
    \twocolumn[\box#2\par\vskip\dbltextfloatsep]%
  \fi}

\newenvironment{delayedtop*}[3]{
  \def\delayedtop@box{#3}
  \def\delayedtop@args{{#2}{#3}}%
  \begin{lrbox}{#3}\begin{minipage}{\textwidth}%
    \def\@captype{#1}%
}{
  \end{minipage}\end{lrbox}%
  \global\setbox\delayedtop@box=\copy\delayedtop@box
  \expandafter\delaytop\delayedtop@args
}

\makeatletter
\renewcommand\paragraph{\@startsection{paragraph}{4}{\z@}%
                                    {1.25ex \@plus0.5ex \@minus.2ex}%
                                    {-1em}%
                                    {\normalfont\normalsize\bfseries}}

\makeatother

\usepackage{silence}
\usepackage{iccv}
\usepackage{times}

\newcommand{\softP}{\mathcal{P}}

\newcommand{\source}{\mathcal{X}}
\newcommand{\target}{\mathcal{Y}}

\makeatother

\usepackage[pagebackref=true,breaklinks=true,colorlinks,bookmarks=false]{hyperref}

\iccvfinalcopy

\def\iccvPaperID{3214} 

\begin{document}

\title{Dual Geometric Graph Network (DG2N) \\ Iterative network for deformable shape alignment}

\author{Dvir Ginzburg\\
Tel Aviv university\\
{\tt\small dvirginzburg@mail.tau.ac.il}
\and
Dan Raviv\\
Tel Aviv University\\

{\tt\small darav@tauex.tau.ac.il}
}


\maketitle


\begin{abstract}
We provide a novel approach for aligning geometric models using a dual graph structure where local features are mapping probabilities.
Alignment of non-rigid structures is one of the most challenging computer vision tasks due to the high number of unknowns needed to model the correspondence. We have seen a leap forward using DNN models in template alignment and functional maps, but those methods fail for inter-class alignment where non-isometric deformations exist. 
Here we propose to rethink this task and use unrolling concepts on a dual graph structure - one for a forward map and one for a backward map, where the features are pulled back matching probabilities from the target into the source.
We report  state of the art results on stretchable domains' alignment in a rapid and stable solution for meshes and point clouds\footnote{Our code will be publicly available upon publication}.
\end{abstract}

\section{Introduction}

\begin{figure}
\begin{center}
\hspace{-26pt}
\setlength\tabcolsep{1.5pt}
\begin{tabular}{>{\footnotesize}cp{1.5cm}}

         \includegraphics[scale=0.55,valign=c]{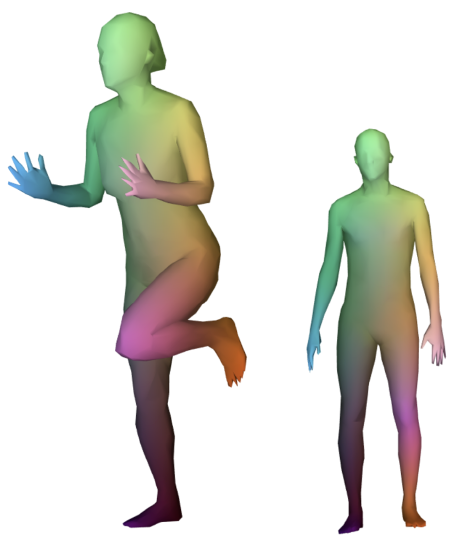} & Non-isometric      
\\

           \includegraphics[scale=0.55,valign=c]{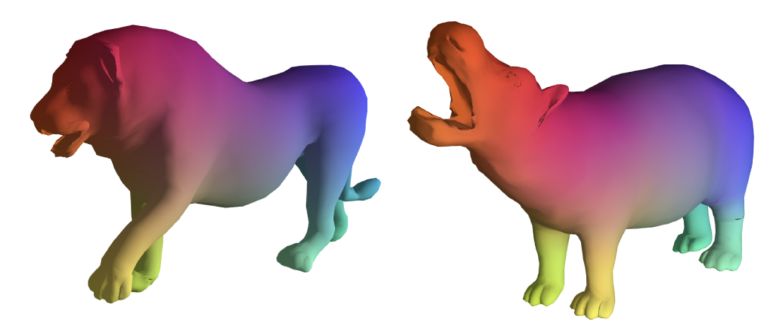}   & Multi-class 
\\

  \includegraphics[scale=0.55,valign=c]{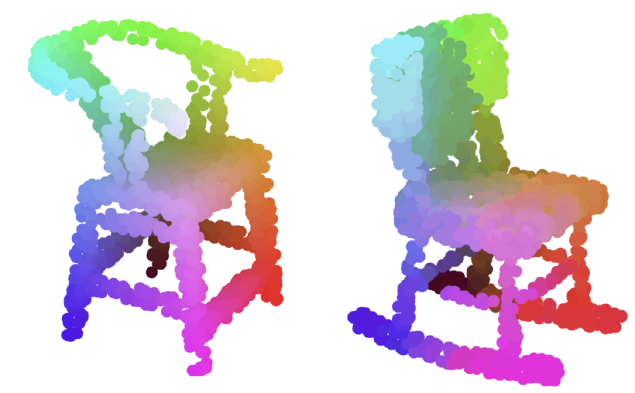} & Various topologies               \\
 
\end{tabular}
\end{center}
\caption{Dense shape correspondence maps generated by DG2N. Similar colors represents correspondence mapping from the source shape (left) to the target (right).} \label{fig:Tizer_pairs}

\end{figure}
The alignment of non-rigid shapes is a fundamental problem in computer vision. It plays an important role in multiple applications such as pose transfer \cite{rad2018feature}, cross-shape texture mapping \cite{zhang2019image}, 3D body scanning \cite{faust}, and simultaneous localization and mapping (SLAM) \cite{wolter2004shape}. 
The task of finding dense correspondence is especially challenging for non-rigid shapes, as the number of variables needed to define the mapping is vast, and local deformations might occur.
To this end, a variety of solutions were offered to solve this problem, using axiomatic and learnable methods. From defining unique key-points or local descriptors and matching such descriptors between the shapes~\cite{hks,wks,gps,shot}, spectral-based methods that try to align the spectra of the shapes \cite{fmnet,surfmnet,halimi,dvir}, or template-based approaches that assume a known pre-defined structure closely resemble all shapes and find the correspondence from each shape to that template \cite{3dcoded}.

Many algorithms for non-rigid alignment relax the problem to matching probabilities which grants them the possibility to consider noise and variability in the pipeline. The transition from soft mapping to vertices alignment, or directly matching points, requires a post-processing step to remove outliers and smooth the results. Unfortunately, this is a slow process and is not performed in a network, as those algorithms are resource-demanding and require a large number of repetitions \cite{vestner2017product,kuhn1955hungarian}.

In this work, we focus on the refinement of a non-rigid alignment map. We unroll the refinement process into a multi-block graph neural network that performs the map denoising. To denoise the alignment in a learnable manner, we construct a dual graph structure, one for the forward map and one for the backward map, where we claim that the features are the actual probabilities for mapping pulled back from the target.
In simple words, what best describes a point in the source, is not just its local features but how do all the points in the target resemble it. We call that structure a \emph{Dual  Geometric Graph Network (DG2N)}.
As the method does not depend on the modality of the input, DG2N is able to refine both meshes and point-clouds under various deformations, as demonstrated in the experimentation section \ref{sec:experiments}.
We report state-of-the-art results on multiple benchmarks and succeed in providing a stable solution even under large non-isometric deformations.

\paragraph{Contributions}
We present three key contributions:
\begin{itemize}
  \item Build a new architecture for self-supervised non-rigid alignment based on a residual pipeline that converges into a clean, soft mapping matrix for each pair of models (zero-shot).
  \item Present a novel concept for graph features derived from the soft alignment map between the shapes.
  \item Report state of the art results in a wide range of benchmarks, including FAUST, TOSCA, SURREAL, SMAL, and SHAPENET.
\end{itemize}
\section{Background}
\vspace{-0.1cm}
This work is focused on alignment refinement of an initial map between non-rigid models , motivated by denoising concepts of graph neural networks.
Let us elaborate on each one of those elements.



\paragraph{Graph neural networks}
While deep learning effectively captures hidden patterns in grid sampled data, we witness an increasing number of applications where the information is better represented in graphs or manifolds \cite{appnp,chen2015signal,dgcnn}. New challenges arise from a non-Euclidean structures due to the variable size of neighbors and unordered nodes.

Graph neural networks go back to 1997, working on acyclic graphs \cite{sperduti1997supervised}, but the notion of graph neural network was officially introduced by Gori \etal in 2005 \cite{gori2005new}.  Within the idea of graph neural networks, the most relevant to this work are convolutional graph neural networks, also known as ConvGNN. Under this umbrella, we can find two main streams; spectral and spatial. The first prominent research on spectral networks was presented by Bruna \etal \cite{bruna2013spectral}. On the other hand, a spatial convolutional structure was addressed more than a decade ago by Micheli \cite{micheli}, which has recently been resurfacing, showing its usefulness for multiple tasks in geometry and computer vision. 

A variety of modern graph learning algorithms \cite{defferrard2017convolutional,gat,appnp,GCN,monti2016geometric,Fey/Lenssen/2019} replaced the traditional Euclidean convolution with a general concept of pulling that can be implemented on a graph.
Among popular modern graph neural network architectures for computer vision tasks, we can find PointNet \cite{pointnet}, its successor PointNet++ \cite{qi2017pointnet} and DGCNN ~\cite{dgcnn}, which provides useful tools to convolve over a set of points. 

In this paper, the unit blocks we use are based on top of the graph convolution network \cite{GCN}, the vertices are points in space, edges are based on the input modality, which is the triangulation for meshes or euclidean nearest-neighbors for point clouds, and the features are alignment probabilities in between the source and the target.


\paragraph{Non-rigid shape correspondence}
Non-rigid shape matching is built out of aligning points with similar features, geometric and/or photometric, and a smoothness term, making sure a point can not be mapped farther from its neighbor.
Under this umbrella, we had seen various axiomatic methods focused on distances, angles, and areas \cite{gps,shot} where a large leap forward was made when deep learning was applied on top of geometric data. 
We can split deep models into two categories - spatial and spectral. Under the spatial approach, we usually see a flow mechanism where the models' changes are minor ~\cite{wu2020pointpwcnet,liu2019flownet3d} or an all-to-all correlation approach \cite{puy2020flot} that can cope with large displacements. When the domain is well defined, then template matching showed great results, as seen in \cite{3dcoded} and in \cite{kanazawaHMR18}.
On the spectral side, various methods based on functional maps \cite{fmnet,dvir,halimi} showed superb results on meshes and points and even excelled on partial alignment \cite{fmnet}. Those papers' goal was to construct deep local features such that the spectra of the shapes would align following a point-to-point soft correspondence  matrix.

One of the challenges in non-rigid alignment is the lack of labeled data. That is mainly because there is no feasible way to own the exact dense correspondence of bendable and stretchable domains on real scanned sets. To overcome this obstacle and remain within the learnable regime, we consider a self-supervised approach.
In the spatial domain, we have seen several useful cost functions that use templates while forcing smoothness on the structures \cite{3dcoded}. More sophisticated assumptions on the domain, such as isometry, were able to learn a mapping by minimizing the Gromov Hausdorff metric as it only needed to compare distances between pairs \cite{halimi}, but failed to converge once stretching appeared. A recent mapping with a cyclic loss measuring the error only on the source showed superior results even under local stretching \cite{dvir}.

All those methods break once there isn't enough data to train. As reported by the authors, either the system can not converge, or we witness a high number of outliers.
To overcome this limitation, we present a zero-shot alignment architecture between two-shapes, where we rethink the alignment process as denoising a soft correspondence matrix. By that, we quickly converge into a clean outlier-free model and can cope with inter-class alignments even under large deformations.


%


\paragraph{Correspondence refinement}
While the methods mentioned in the previous section brought for the first time the capability to densely align between 3D objects with satisfying results, still, most output maps were noisy, partial, or sparse.
For the extent of our knowledge, we are the first to offer a learnable refinement pipeline, 
nevertheless, many axiomatic methods have tried to iteratively refine and sharpen these maps.
Most refinement methods solve the \textit{optimal transport problem} between the shapes under various constraints derived from the input mapping. One example of such approach was even presented in the original functional maps paper, where the authors showed how ICP \cite{icp} applied to the spectral features improves the map dramatically.
Others \cite{mandad2017variance} solve the transport problem directly under constraints as geodesic distance preservation or variance minimization of the mapping.
Methods as the Product Manifold Filter (PMF) \cite{pmf}  use linear or quadratic assignment solvers \cite{kuhn1955hungarian} to determine the solution to the transport problem.
Lately, an important method named ZoomOut \cite{zoomout} showed how one can refine the initial map in the spectral domain by progressively increasing the dimension of the functional mapping, refining at each step.
While the above axiomatic methods are milestones in the field, methods that try to solve the transport problem in the spatial space become computationally unfeasible and slow even under relatively sparse input sampling. On the other hand, spectral methods solve the alignment problem with a decent outcome for low-frequency maps but struggle with high frequencies. Furthermore, the functional maps setting used in methods like ZoomOut assume isometry between shapes,  thus present degraded results in local scaled and deformable shape matching.
Adding to the above, all spectral methods are based on the Laplace Beltrami operator which is known to be unstable and generate poor spectra under modalities different that meshes, like point clouds. Due to that, such methods tend to show inconsistent results, or even harm the initial map in some settings, as we present in the results section \ref{sec:experiments}.


\paragraph{Graph denoising}
Denoising graph signals is a ubiquitous problem that plays an important role in many areas of machine learning \cite{appnp,graphdenapp1,graphdenapp2}, and was proven to improve results on a wide range of problems \cite{graphden1,graphden2}.
The two main approaches for analyzing graph signals are graph regularization-based optimization and graph dictionary design \cite{graphdict1,graphdict2}; 
The optimization approach applies a regularization term that promotes certain characteristics on the model, such as smoothness or sparsity \cite{chung1997spectral,graphden1}. 
The optimization function itself usually takes the form of 
$$
\argmin_{\bm{x}} ||t-x||_2^2 + \lambda Q(x)
$$
where $t$ is the noisy graph signal, and $Q$ is the regularization term.

When smoothness of the graph signal is assumed, one popular choice is the quadratic
form of the graph Laplacian, the Dirichlet energy, discretized as $x^T\mathcal{L}x$ where $x$ is the graph signal, which captures the second-order difference of a graph signal \cite{moon2000mathematical}. For sparsity of the graph signals, a graph total variation term that captures the first-order difference of the graph signals was proven effective \cite{graphdenapp1,chen2015signal}. Recently, denoising graphs using deep architecture showed superior results by unrolling the $L_1$ regularization term into several layers, converging iteratively into the desired cost function \cite{chen2020graph}.


\section{Method}
\begin{figure*}[t!]
  \centering
  
  \includegraphics[width=\linewidth]{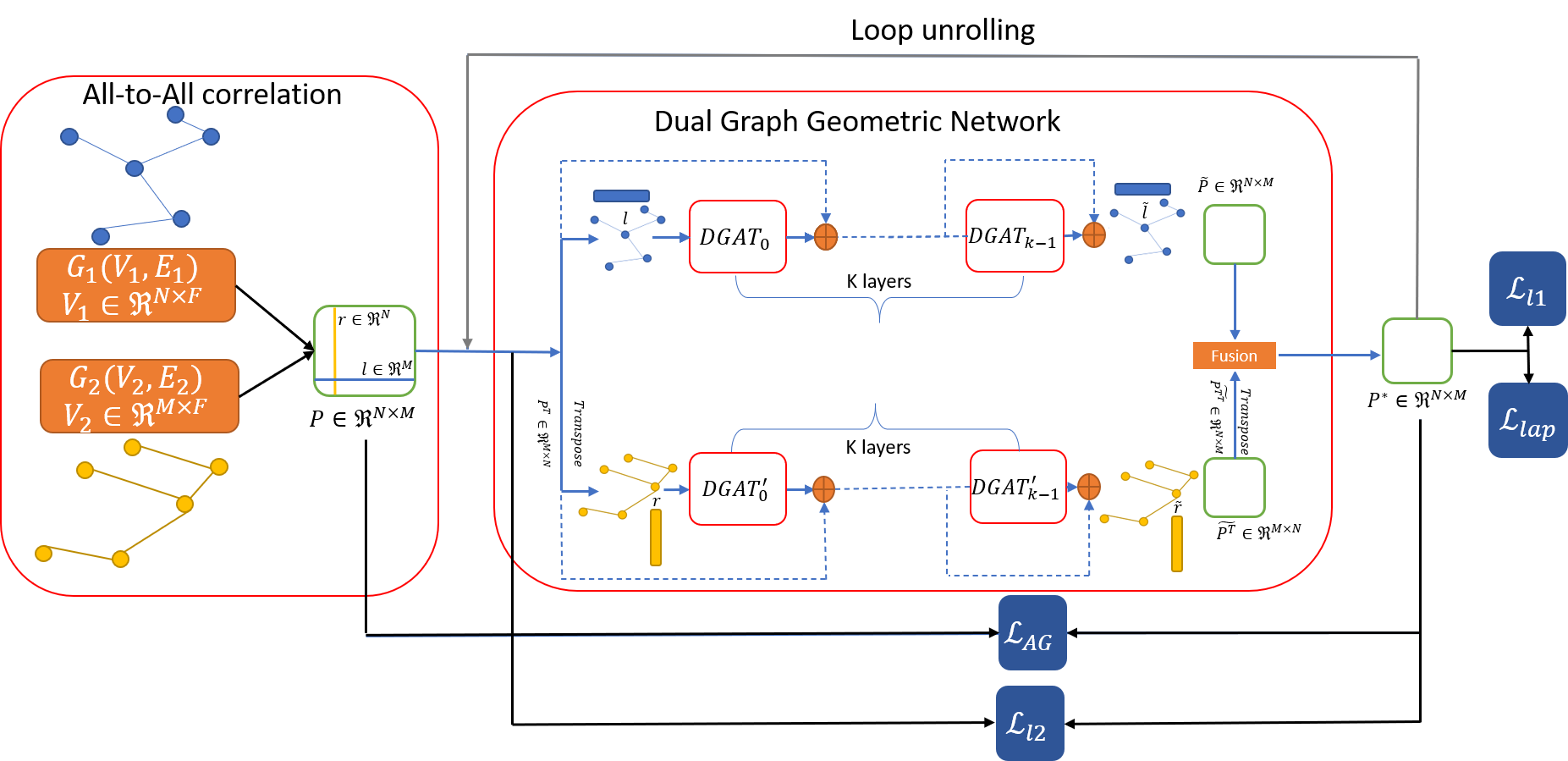}
  \caption{Our dual graph geometric network. Given two input graphs we first pass them through  an initiator to have the initial soft correspondence matrix $\softP$, representing correspondence probabilities between all vertex-pairs in the mapping. We then pass the graphs induced by $\softP$ and $\softP^T$ through stacked layers of $DGAT$ with residual connections, where the output is a refined soft correspondence matrix. The 4 loss objectives (Section \ref{subsec:losses}) allows iterative refinement over 
$\softP$, where the input $\softP$ for the next iteration is the output of the previous one.}
  \label{fig:architecture_scheme}
\end{figure*}

The proposed method is a self-supervised learnable pipeline. To align two non-rigid models, we use point embeddings generated by a black box model we refer as the initiator.
Using these embeddings we define the soft-alignment map as the cosine similarity between the source and target embeddings.
We learn how to iteratively update the soft correspondence matrix to improve the results and remove outliers \ref{subsec:dg2n}. 
Our cost function is based on the understanding that each point's best features are the correspondence probabilities to the target points.
Specifically, if $P$ is a soft correspondence matrix, i.e., if $P \in \mathbb{R}^{N \times M}$, and $P_{ij} \in [0,1]$ is the probability that point  $i$ matches point $j$,  then in the primal graph, the features of point $i$ are the $i$'th row of P, and the features of the dual graph are represented by the columns (or rows of $P^T$). We denote this primal-dual structure as the Dual Graph Geometric Network (DG2N). 

While we use the cosine-based soft correspondence map as $\softP$, DG2N can work with any soft-alignment matrix, such as the soft-alignment map proposed in \cite{fmnet}, and extended in \cite{halimi,dvir}. 

In the scenario of point clouds, where there is no relevant dense-correspondence initiator (Section \ref{subsec:point_clouds}), we use a rigid-alignment algorithm such as DCP \cite{dcp} as our feature extraction network. This is an extremely weak learner that produces outliers and inaccurate alignments, but it is sufficient to train the proposed DG2N architecture and converge to a very good mapping.

DG2N is composed of our new graph attention mechanism, activated on the two graphs (primal and dual) simultaneously. We refer to the new convolution blocks by differential-GAT, or DGAT. Inspired by \cite{gat}, we consider a pulling strategy in-between points and their neighbors, where we concatenate the differences between the node features for a fixed number of neighbors. 

To keep improving the outcome and not collapsing during the denoising process, we present four cost functions. We require the alignment to be injective, smooth, not too far from the previous iteration, and to keep the most valuable points in place.

In what follows, we elaborate on the main three components of the architecture's pipeline and the four cost functions.

\subsection{Architecture}
\paragraph{All-to-All mapping}\label{subsec:softcorr}
To achieve a coherent and smooth correspondence map between two shapes, our dual graph unit (DG2N) uses the soft correspondence mapping $\softP$ as an input.
We can use any known method which has a soft correspondence matrix in the pipeline as an initiator, for example \cite{dvir,fmnet,halimi}.\\
In detail, \cite{fmnet} showed that using the functional mapping $\mathcal{C}$, with the graphs laplacian eigendecomposition of the shapes $\Phi,\Psi$ the soft correspondence is constructed by 
\begin{equation}\label{eq:PfromFM}
    \softP \propto |\Psi \mathcal{C} \Phi^T|.
\end{equation}
In the scenarios where spectral methods are unstable or fail to create reasonable results (Section \ref{subsec:animals}) we show here that an elementary all-to-all correlation matrix can be constructed from popular rigid alignment networks such as DCP \cite{dcp}. We found that to be good enough as an initiator for the refinement.
Specifically, we use the last hidden layer of DCP as a point descriptor $h_{x_i}$.
The soft correspondence is constructed by the cosine similarity between the descriptors:
\begin{equation}
        \softP_{i,j}=\frac{ h_{x_i} \cdot h_{x_j}} {|| h_{x_i} ||_2 \cdot ||h_{x_j}||_2},
\end{equation}

where $h_{x_i}$ and $h_{x_j}$ represent two feature vectors of points $x_i$ and $x_j$.

In Section  \ref{sec:experiments} we show such a simple solution finds a noisy correspondence  between non-isometric pairs but provides sufficient initialization for our architecture.

\paragraph{Differentiable GAT}\label{subsec:DGAT}
\begin{figure}[t]
  
  \includegraphics[width=\linewidth]{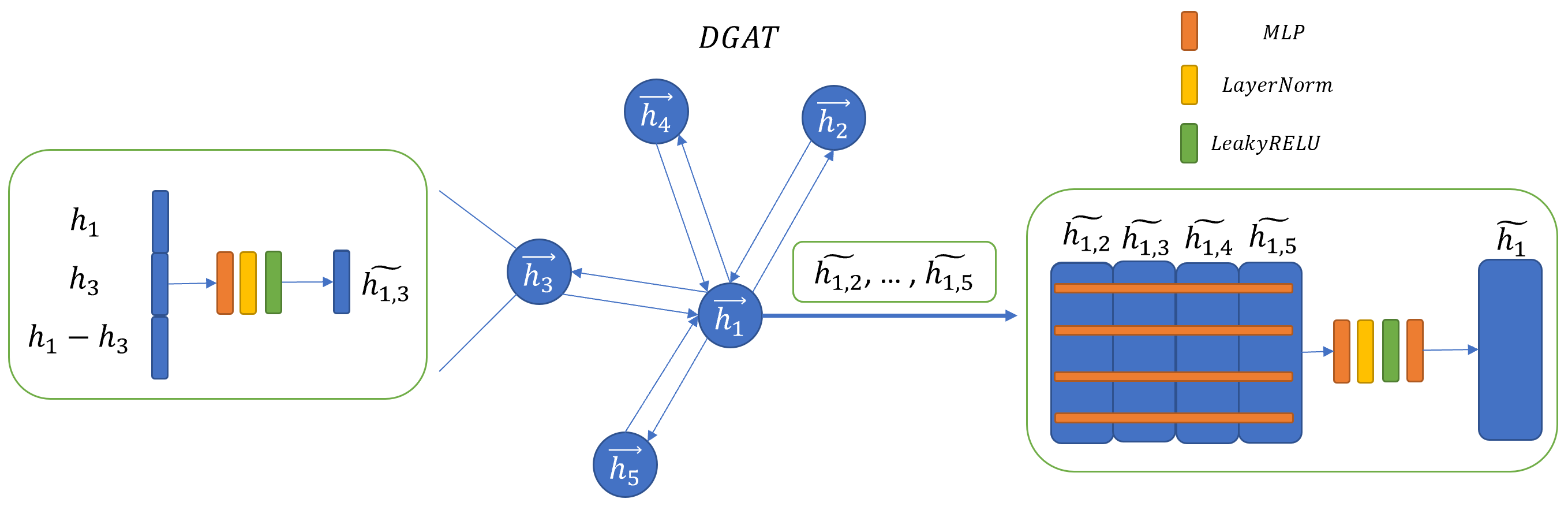}
  \caption{Single DGAT layer.  Phase I: Create the \textit{difference guiding feature vector} $[\vec{h_i}|\vec{h_j}|\vec{h_i}-\vec{h_j}]\in \mathbb{R}^{3M}$ and pass it through a first learnable architecture.   Phase II: stack phase I features $\tilde{H_i}\in\mathbb{R}^{M\times K}$ and regress each feature individually through the second module forming the output feature per point $\tilde{h_i}\in\mathbb{R}^{M}$.}
  \label{fig:gdcarch}
\end{figure}

DG2N is composed of two parallel GNN modules based on the Differential Graph Attention (DGAT) layer. Inspired by GAT \cite{gat}, DGAT perform weighted local pooling, only here we stack the output feature vector from the per-pair stage and apply a per-feature network to learn the best refinement step.

The most generalized structure associated with GNNs is
$$
f_{i}^{'}=\gamma_{\Theta}(f_i,\msquare_{j\in \mathcal{N}(i)}\phi_{\Theta}(\psi(f_i,f_j,e_{j,i})))
$$
where $f$ and $f^{'}$ represent the input and output data channels respectively, $\msquare$ is some differentiable aggregation function, $\gamma_{\Theta},\phi_{\Theta}$ denote non-linear transmission functions, and $\psi$ defines the feature-fusion method applied before the features propagate through $\phi_{\Theta}$.

$DGAT$ set $\msquare$ to be stacking of the per-neighbor output feature vector, $\psi$ is the guiding vector function 
\begin{equation}
    \psi(f_i,f_j,e_{j,i}) = [f_i|f_j|f_i-f_j],
\end{equation}
and the graph's edges determine the neighborhood. $\gamma_{\Theta}$, and $\phi_{\Theta}$ are variants of a multi-layer perception (MLP) with normalization and non-linear activation function layers.
In practice, DGAT takes the form
\begin{equation}\label{eq:GDC}
    f_{i}^{'}=DNN_2(\underset{{j\in \mathcal{N}(i)}}{\mathbin\Vert}(DNN_1([f_i|f_j|f_i-f_j]))
\end{equation}
where $\underset{{j\in \mathcal{N}(i)}}{\mathbin\Vert}$ is the concatenation of $DNN_1$ outputs.
An illustration can be found in Figure \ref{fig:gdcarch}.\\ \\ 
One crucial emphasis here is the role of $DNN_2$ and the distinction to other suggested aggregation functions. One optional aggregation would be to concatenate the output difference features resulting in a feature vector of dimension $\tilde{H_i}\in\mathbb{R}^{KM}$, while we stack the features resulting in $\tilde{H_i}\in\mathbb{R}^{M\times K}$.
Our construction not only produces a learnable module with a factor of $M^2$ fewer parameters as $DNN2$ is applied on the $k$ dimensional vectors but, more significantly,  acts as a learnable weighting function that incorporates the various per-neighbor refinements into a refinement step per-node.
\paragraph{Dual Geometric Graph Network} \label{subsec:dg2n}
In the heart of the proposed architecture, is our understating that the soft correspondence matrix  $\softP$ induces a graph. The nodes of the \emph{primal} graph are the source points, and the features are the correspondence measure for all target points, i.e., the rows of $\softP$ are the features.
The \emph{dual} graph has the same structure only based on $\softP^T$. Here the nodes are the vertices of the target shape, and the columns of $\softP$ are the features.
We provide a visualization of the architecture in 
Figure \ref{fig:architecture_scheme}.
Each primal-dual pair  ($\softP,\softP^T$) is passed through $k$ layers of $DGAT$  in a res-net structure; i.e. the output of each $DGAT$ layer is $DGAT(\softP) + \softP$ for an input soft correspondence matrix $\softP$, and similar for $\softP^T$, the dual graph pipeline.
In each iteration, we fuse $\softP$ and $\softP^T$ into one aligned soft correspondence matrix\footnote{There are several reasonable options for the fusion, as element-wise max or mean. In practice, no consistent improvement was noted by one option over the other.}.\\ The output of each iteration of DG2N refinement is also a soft correspondence matrix. As the correspondence statistics vary between one iteration to the next, we use different weights per iteration. As we continue to iterate, the soft correspondence matrix improves and converges to a clean, outlier-free soft mapping. At inference, the output map is the maximum-likelihood solution derived from the soft correspondence matrix, which is:
\begin{equation}\label{eq:hard_map}
    \pi(\mathcal{X})_i=\argmax_j \softP_{ij}.
\end{equation}
Where $\mathcal{X},\mathcal{Y}$ are the source and target shapes respectively.

\subsection{Losses}\label{subsec:losses}
We combine four different losses in this pipeline. $\mathcal{L}_{L}$ Laplacian loss, $\mathcal{L}_{l1}$ Sparsity loss, $\mathcal{L}_{AG}$ Anchors guidance loss
and $\mathcal{L}_{l2}$ Denoising regularization.

These constraints form together the loss objective of a single refinement step of DG2N, which is:
\begin{equation}\label{eq:graphlosswithssar}
\begin{aligned}
    \mathcal{L}=\mathcal{L}_{L} +  \mathcal{L}_{l1} + \mathcal{L}_{AG} + \mathcal{L}_{l2}
\end{aligned}
\end{equation}
All four losses are evaluated separately and summed together both for the primal and dual graphs and executed for every iteration output.

Let us elaborate on each term in the loss.

\paragraph{Laplacian loss}
  Laplacian regularization term pushes toward graph smoothness. 
  It takes the form of:
  $$\mathcal{L}_L= \lambda_{L} \softP^T L_\softP \softP = \lambda_{L} \sum_{(i,j) \in E_{\source}}w_{i,j}||\softP_{i,\star}-\softP_{j,\star}||_2^2
$$
 where $L_\softP=D-A$ is the graph Laplacian of the source shape $\source$, $D$ is the degree of each node, $A$ is its adjacency matrix, and $\softP_{i,\star}$ is the $i$'th row of $\softP$. 
 Two important items to note here, the first is that this term can be used on any structure inducing a graph. Second, while all other methods use Laplacians on the shape coordinates in space, we claim that smoothness should apply directly to the soft correspondence matrix. Since the features are the mapping probabilities, we claim the smoothness on $\softP$ is a better goal.
 
\paragraph{Sparsity regularization}
We add the $\mathcal{L}_{1}$ regularization on the rows of $\softP$. Specifically,
\begin{equation}
    \mathcal{L}_{1} = \lambda_{l1} \sum_{i=1}^N|\softP_{i,\star}|_1.
\end{equation}
As the rows of $\softP$ represent the alignment probabilities, we wish to promote sparsity. Each source point corresponds to a single target point, meaning one element should hold most of the energy, and the rest should decline rapidly.
Note that we are not normalizing the rows or columns in each iteration, thus $\softP$ is in fact a pseudo probability matrix, where each row does not guaranteed to sum to one.

\paragraph{Anchors guidance loss}\label{subsec:anchor_guidence}
One of the caveats with the Laplacian regularization is its tendency for over-smoothing and thus hurt the overall performance \cite{wu2019simplifying}.
In our case, this phenomenon takes the shape of pushing all correspondence probabilities of $\softP$ towards the average. While this decreases the Laplacian loss, it results in significant degradation of the results, as shown in the ablation study (Table \ref{tab:Ablation}).

To solve the mentioned problem, we present a self-supervised anchor guidance mechanism.
Motivated by node classification tasks \cite{giles1998citeseer,sen2008collective},
a few anchor points are sampled from the initial soft correspondence map, and we seek to use their initial mapping as guidance through the refinement.
Bear in mind that those points are not fixed and are part of the learnable pipeline, only we provide extra attention to points we believe in their mapping.

Analyzing soft correspondence mappings generated from different pipelines (FMnet \cite{fmnet}, SURFMNet \cite{surfmnet}, DCP \cite{dcp}), we observed two attributes that reoccur by all algorithms:
\begin{enumerate}
    \item Source nodes where the highest correspondence probability of $\softP$ is two orders of magnitude larger than the average probability ($\frac{1}{M}$) usually point to the true correspondence.
    \item High probability correspondences reside in clusters, that is, if a source node corresponds to some target node with high probability, it is usually the case its neighbors will also have high probability correspondence to some neighbor of this corresponding point.

\end{enumerate}

We utilize the above observations to attend the over-smoothing caused by the Laplacian.
For each $\source,\softP,\target$ we first sample $k\leq|V_{\source}|$ disconnected nodes using FPS \cite{FPS} 
noted as $V_{K_{\source}}$, and assign their soft label by defining 
\begin{equation}\label{eq:pernodeSSAG}
\begin{aligned}
\hat{y}_{i}=\argmax_j\softP&_{i,*} \quad \forall v_i \in V_{K_{\source}} \\
C(\hat{y}_{i}) = \softP&_{i,\hat{y}_{i}}
\end{aligned}
\end{equation}
where $C(\cdot)$ is the confidence $x_i$ corresponds to $\hat{y}_{i}$.

Using the above formulation, we define a soft-classification problem. We constrain the network to label the anchor points similarly to how they were classified before the refinement layer. The penalty for each wrong classification is directly proportional to the confidence $C(\cdot)$. 

  We use the anchors' notations and define the anchor loss as the cross-entropy between the presumed label, and the output features of DG2N Layer. Specifically,
\begin{align}
 \mathcal{L}_{AG}=\sum_{x_{i}\in V_{K_\source}} &C(\hat{y}_{i})^2 \\ \nonumber &\Big(-x_i[\hat{y}_{i}] + \log\big(\sum_{j=0}^{|V_{K_\source|-1}}\exp{(x_i[j])}\big)\Big),  \label{eq:AG_los}
\end{align}
where $x_i[m]$ is the $m$'th element in the row of $\softP$ corresponding to vertex $x_i$.

\paragraph{Denoising regularization}
For each iteration, we assume the output is similar to the input. By that we force the network to penalize for large gaps, and de-facto promote minor updates, usually referred to as noise or outliers in this paper. 
Denoting the previous layer matrix as $\softP$ and the output of a single DG2N iteration as $\softP^*$, the denoising regularization takes the form of:
\begin{equation}\label{loss:l2}
    \mathcal{L}_{l2}=\lambda_{l2} ||\softP^{*}-\softP||_2^2.
\end{equation}

\section{Experiments}\label{sec:experiments}
The following section presents multiple scenarios in which our self-supervised architecture surpasses current state-of-the-art algorithms for non-rigid alignment.
In addition, we will present our zero-shot pipeline that achieves near-perfect results for non-isometric deformable shape matching.
To adjust DCP \cite{dcp} for the dense correspondence task we follow FMnet \cite{fmnet} loss and optimize for $\mathcal{L}=||P-I||_2^2$, as during DCP training the map is given by ~$\pi(\mathcal{X}_i)=\mathcal{Y}_i$. 

\begin{table*}[t]
\begin{tabular}{l|cccccc}
 & FAUST \cite{faust}        & SURREAL \cite{surreal}      & F on S       & S on F       & SMAL \cite{smal} & SMAL on TOSCA \cite{tosca} \\ \hline
FMNet \cite{fmnet}                                      & 12.1         & 18.7         & 35.3         & 33.4         & *    & *             \\
3D-CODED \cite{3dcoded}                                   & 8.5          & 15.5         & 28.5         & 26.0         & 8.8  & 35.7          \\
Deep GeoFM \cite{deepfm}                                 & 3.8          & 4.2          & 7.8          & 14.2         & *    & *             \\ \hline
DCP(Unsup) \cite{dcp}                            & 19.3          & 21.2         & 26.4         & 28.9         & 16.8    & 27.3            \\
SURFMNet(Unsup) \cite{surfmnet}                            & 7.1          & 11.3         & 31.5         & 42.3         & *    & *             \\
Unsup FMNet(Unsup) \cite{halimi} 
        & 13.1         & 14.6         & 33.2         & 38.5         & *    & *             \\
\hline
PMF \cite{pmf} on DCP                          & 18.1          & 19.8         & 21.6         & 25.8         & 14.0    & 23.9            \\
PMF \cite{pmf} on GeoFM                           & 3.5          & 4.1         & 6.6         & 8.5         & *    & *            \\
ZoomOut \cite{zoomout} on DCP                          & 22.9          & 18.5         & 28.0         & 29.9         & 14.6    & 26.8            \\
ZoomOut \cite{zoomout} on GeoFM                           & \textbf{2.9}          & \textbf{3.8}         & 6.3         & 8.4         & *    & *            \\
\hline
Ours(Unsup) on DCP \cite{dcp}                       & 15.3         & 12.9         & 21.1         & 25.4         & \textbf{7.9}    & \textbf{19.5}             \\
Ours(Unsup) on FMNet \cite{halimi}                       & 9.6         & 11.8         & 13.5         & 14.9         & *    & *             \\
Ours(Unsup) on SURFMNet \cite{surfmnet}                            & 5.9          & 8.1         & 9.3         & 10.5         & *    & *             \\
Ours(Unsup) on GeoFM \cite{deepfm}               & 3.4 & 4.1 & \textbf{6.2} & \textbf{8.1} & *    & * \\ \hline           
\end{tabular}
\caption{Mean geodesic error (MGE) comparison by different methods on FAUST(F), SURREAL(S), SMAL and TOSCA datasets. No post processing filters are used for any of the methods except the specified refinement procedures. We remark that due to the numerical instabilities of the Laplacian decomposition we were not able to run the spectral methods with the code published by the authors (results marked with *) on our re-sampled SMAL dataset.
ZoomOut presents slightly better results than DG2N for refining GeoFM, where GeoFM already achieves remarkable results on FAUST, without any post-processing filters. However, when refining imperfect initiators, such as DCP, ZoomOut compromises the initiators' results, \textit{degrading} the MGE by 3.6 points. In contrast, our DG2N is robust to imperfect initiators, and \textit{improves} the noisy correspondence of DCP by 4.0 points.}
\label{tb:fausursmal}
\end{table*}


We evaluate DG2N on a wide range of popular datasets for dense shape correspondence. To assess the network's robustness, we test it on multiple datasets with different statistical and topological attributes as humans datasets  (FAUST\cite{faust} and SURREAL \cite{surreal}), animals (SMAL \cite{smal} and TOSCA \cite{tosca}) or chairs and plains (SHAPENET\cite{shapenet}). We use a remeshed and down-sampled version of FAUST, SURREAL, and SMAL, as suggested by \cite{ren}. In the generated datasets, each shape has approximately 1000 vertices. These re-meshed datasets offer significantly more variability in terms of shape structures and connectivity than the original datasets \cite{deepfm}.

\paragraph{Mesh Error Evaluation}
The measure of error for the correspondence mapping between two shapes will be according to the Princeton benchmark \cite{geodesic_error_metric}, that is, given a mapping $\pi_{\to}(\mathcal{X, Y})$ and the ground truth $\pi^*_{\to}(\mathcal{X, Y})$, the error of the correspondence matrix is the sum of geodesic distances between the mappings for each point in the source figure, divided by the area of the target figure.
\begin{equation} \label{eq:geodesic_error}
\epsilon(\pi_{\to}) = \sum_{x \in \mathcal{X}} \frac{\mathcal{D_Y}(\pi_{\to}(x),\pi^*_{\to}(x))}{\sqrt{area(\mathcal{Y})}},
\end{equation}
where the approximation of $area(\mathcal{\cdot})$ for a triangular mesh is the sum of its triangles area.

\paragraph{Humans datasets - FAUST and SURREAL}\label{subsec:faust_scape}
We follow the suggested setting \cite{deepfm} for these human datasets and split both datasets into training sets (80 shapes) and test sets (20 shapes). The specific shape splits are identical for all tested methods for a fair comparison. We test two scenarios, one in which we train and evaluate on the same dataset and one in which we test on the other dataset (e.g., training on FAUST evaluating on SURREAL). This experiment aims
attesting the generalization power of all methods to small
re-meshed datasets, as well as their ability to adapt to a different dataset at test time.

Table \ref{tb:fausursmal} stresses some of the key advantages of DG2N compared to other self-supervision methods and refinement techniques, as robustness and generalization. While almost all learnable methods perform reasonably well on the same-dataset benchmark, we see significant performance gaps  compared to other methods when conducting the cross-dataset test; this is due to the fact we are self-supervised and shape-pair specific, thus are almost invariant to noise added to the system by changing the statistical attributes of the data. 
Comparing to PMF \cite{pmf}, having an average refinement time of two minutes per shape-pair, we see that DG2N outperforms its results in both settings while refining in two orders of magnitude faster.
PMF is extremely sensitive to the optimization hyper-parameters, and in our experiments we observed variability of up to 5X in the MGE score under different parameters. We report here the results under the optimal parameters.
Compared to ZoomOut refining GeoFMNet, which receives remarkable results without any post processing filters, ZoomOut presents 0.5 cm MGE less in the same-dataset setting, and worst results compared to DG2N in all other experiments.
When ZoomOut is examined on noisy initial maps, as in the case of using unsupervised initiators, ZoomOut presents results that are often worse than not using filters at all. We attribute that phenomena to the use of the hard mapping $\pi(\mathcal{X})$ by ZoomOut. $\pi(\mathcal{X})$ of noisy initiators often include cases where neighbor points map to geodesically distant target vertices, using such outliers as the initial conditions for the refinement process may cause divergence, as happened in our evaluations.
Unlike ZoomOut, DG2N takes advantage of the soft alignment matrix, which indicates the pipeline for possible outliers, and offer other mappings that are coherent to the point neighborhood.

\paragraph{Animals datasets - SMAL and TOSCA}\label{subsec:animals}

To better understand the different models' generalization capabilities and ensure the models are not hand-crafted for human-like structures, we also assess the network's performance on animal datasets.
SMAL \cite{smal} dataset provides a generative model
for synthetic animals creation in different categories as cats, horses, etc.; SMAL is extracted from a continuous parametric space with a fixed number of vertices and same triangulation for all shapes, with the possibility of generating "infinitely many" training samples. Unlike SMAL, TOSCA \cite{tosca} contains a fixed selection of shapes, including 9 cats, 11 dogs, 3 wolves, etc. Which is both dramatically smaller and has no topological guarantees, meaning no two shapes have the same triangulation.
The animals datasets experiment was conducted as follows: For each SMAL category, we create 80 shapes for training and 20 for the test, resulting in 500 samples. We must emphasize that previous methods \cite{3dcoded} that worked with SMAL used two orders of magnitude more training samples in their experiments.
Table \ref{tb:fausursmal} expresses the advantages of DG2N over previous works that are considered state-of-the-art in this regime. The tested spectral based methods (FMnet variant \cite{fmnet,halimi}) failed to converge on the remeshed datasets, probably due to the unstable and noise process of the decomposition of the Laplacians. Compared to ZoomOut and PMF we see similar trains to the FAUST and SCAPE experiments, where both achieve an average of 4 cm MGE worse results than DG2N.
\paragraph{Deformable irregular correspondences - point clouds registration}\label{subsec:point_clouds}
Point cloud registration is undoubtedly one of the hardest registration tasks for 3D shapes, while it is the most common scenario in real-world cases. We evaluate the different methods of chosen classes from SHAPENET \cite{shapenet}, namely chairs, cars, and plains; Each category contains multiple subjects, where no pair is isometric, nor has the same number of points. Unlike meshes, point clouds suffer from noise and topology ambiguity due to the sampling process involved in generating them and the surface-approximation heuristics needed to define each point's neighborhood.
Spectral based methods undergo significantly 
degradation in the results, since spectral decomposition of point clouds is inaccurate and unstable ~\cite{pclap1,pclap2}. GeoFMNet is a supervised pipeline, thus irrelevant for this evaluation as ShapeNet doesn't contain any correspondence labels. The authors of SURFMNet did not evaluate on point-clouds nor offered the tools for such evaluation. For fairness, we used \cite{Sharp:2020:LNT} which is a new tool for Laplacian decomposition on point clouds for the evaluation of SURFMNet.
No dataset currently exists with ground-truth correspondences between deformable point clouds, so we turn to evaluate the performance of the different methods visually, in terms of smoothness, coherence\footnote{A good alignment will map a guitar neck of one shape to the other.}, and robustness to deformations. 

\begin{figure}[t]
\centering 
  \begin{subfigure}{0.23\textwidth}
    \centering
  \includegraphics[width=1.5cm]{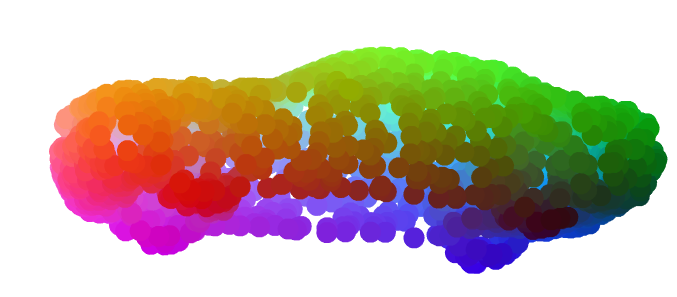}
    \label{fig:1}
  \end{subfigure}\hfil
    \begin{subfigure}{0.23\textwidth}
    \centering
  \includegraphics[width=1.5cm]{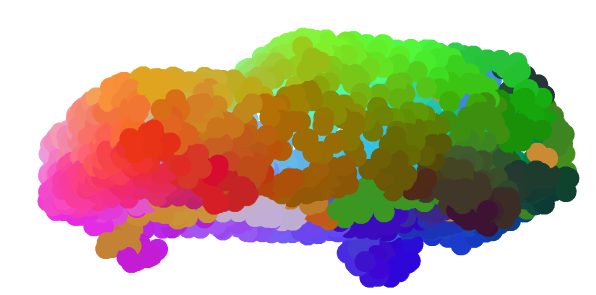}
    \label{fig:1}
  \end{subfigure}\hfil
    \begin{subfigure}{0.23\textwidth}
    \centering
  \includegraphics[width=1.5cm]{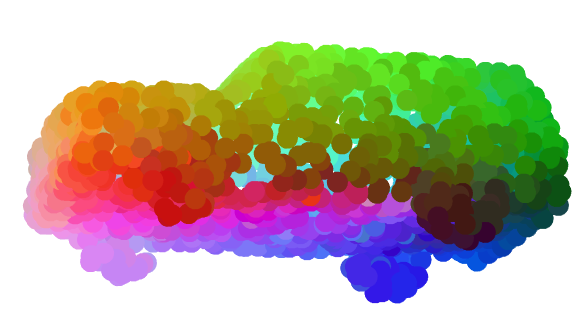}
    \label{fig:1}
  \end{subfigure}\hfil
    \begin{subfigure}{0.23\textwidth}
    \centering
  \includegraphics[width=1.5cm]{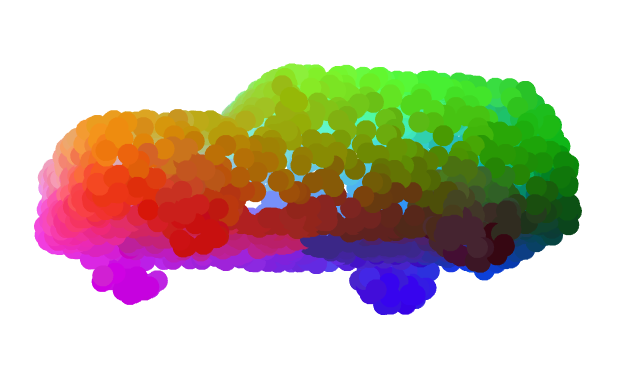}
    \label{fig:1}
  \end{subfigure}\hfil

  \medskip
  
  \begin{subfigure}{0.23\textwidth}
    \centering
  \includegraphics[width=1.5cm]{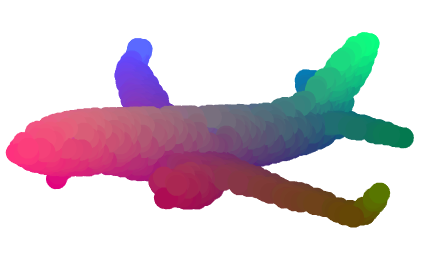}
    \label{fig:1}
  \end{subfigure}\hfil
    \begin{subfigure}{0.23\textwidth}
    \centering
  \includegraphics[width=1.5cm]{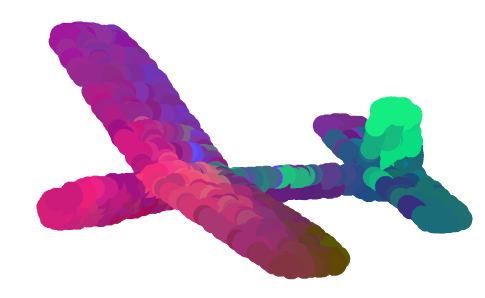}
    \label{fig:1}
  \end{subfigure}\hfil
    \begin{subfigure}{0.23\textwidth}
    \centering
  \includegraphics[width=1.5cm]{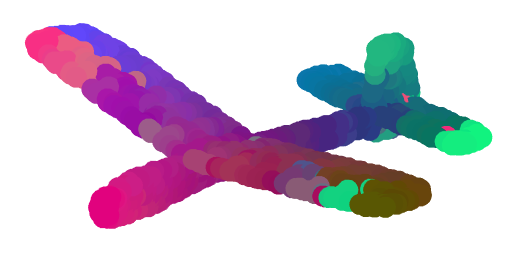}
    \label{fig:1}
  \end{subfigure}\hfil
    \begin{subfigure}{0.23\textwidth}
    \centering
  \includegraphics[width=1.5cm]{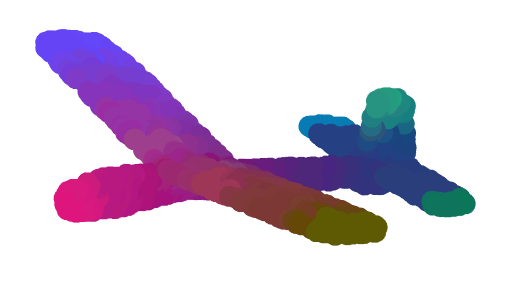}
    \label{fig:1}
  \end{subfigure}\hfil

  \medskip
  
  \begin{subfigure}{0.23\textwidth}
    \centering
  \includegraphics[width=1.20cm]{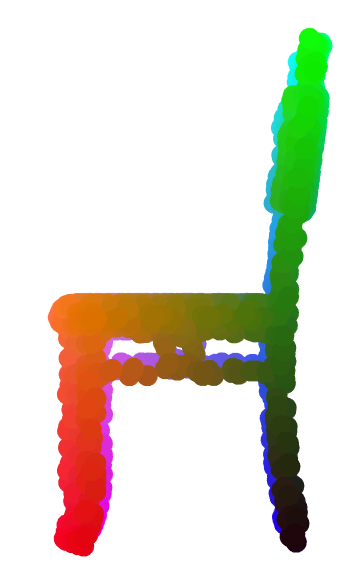}
    \caption{Source\\\ }
    \label{fig:1}
  \end{subfigure}\hfil
    \begin{subfigure}{0.23\textwidth}
    \centering
  \includegraphics[width=1.0cm]{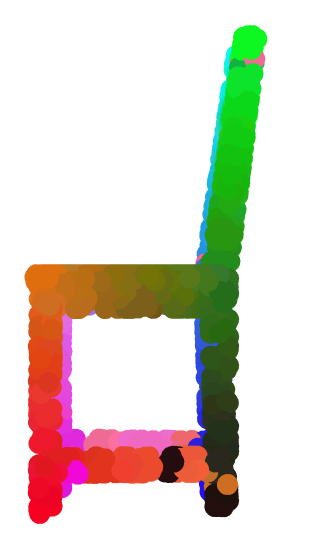}
    \caption{SURFMNet\\*}
    \label{fig:1}
  \end{subfigure}\hfil
    \begin{subfigure}{0.23\textwidth}
    \centering
  \includegraphics[width=1.0cm]{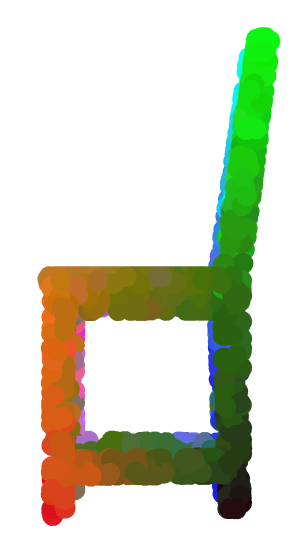}
    \caption{Elementary\\ structures}
    \label{fig:1}
  \end{subfigure}\hfil
    \begin{subfigure}{0.23\textwidth}
    \centering
  \includegraphics[width=0.90cm]{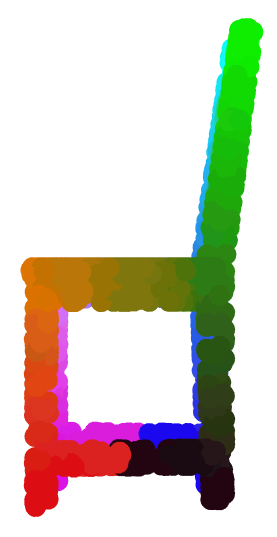}
  \caption{Ours\\\ }
    \label{fig:1}
  \end{subfigure}\hfil
  
  \medskip
  
  \caption{Dense correspondence on point clouds - spectral solutions fail to create smooth or coherent results due to the noisy nature of the Laplacian for point clouds, Elementary structures \cite{deprelle2019learning} and other reconstruction based methods do not enforce smoothness resulting in noisy maps.}
  \label{fig:pc_reg}
\end{figure}
\paragraph{Zero-shot correspondence}\label{subsec:zeroshot}
"Zero-shot" self-supervised methods are essential and brought great achievements and new capabilities in other domains as super-resolution and image generation \cite{zeroshot2,zeroshot1}.
Having a zero-shot registration method for 3D shapes is considered exceptionally difficult, with only a few \cite{dvir,halimi} that tried to tackle the problem. Unfortunately, as seen in previous experiments (Section \ref{subsec:animals}), spectral methods are sensitive and limited in terms of the input domain. To present our self-supervision capabilities, we chose randomly 10  inter-class shape pairs for the FAUST-remeshed dataset. For each pair, we trained DCP \cite{dcp} only on the inference pair until convergence exposing new linear augmentations each training step and ran the inference. On the provided output we ran DG2N refinement scheme. Naturally, only unsupervised methods are relevant for comparison. We present a comparison to other zero-shot methods in Table \ref{tb:self_sup}.

\begin{table}[t]
\begin{tabular}{l|c}
\multicolumn{1}{c|}{Method} & Mean geodesic error \\ \hline
SURFMNet(Unsup) \cite{surfmnet}                                    & 36.2                \\
Unsup FMNet(Unsup)\cite{halimi}                                  & 16.5                \\
Cyclic-FMnet(Unsup) \cite{dvir}                                & 14.1                \\ \hline
DCP(Unsup) \cite{dcp}                                     & 19.5                \\
Ours(Unsup) on DCP                             & \textbf{11.0}                 \\ \hline
\end{tabular}
\caption{Mean geodesic error in a zero-shot setting on FAUST-remeshed. We present best results among all unsupervised methods that are relevant to this experiment setting.}
\label{tb:self_sup}
\end{table}

\subsection{Ablation}
DG2N training is constructed of 4 different loss functions, each plays an important and substantial role in the refinement process.
We provide table \ref{tab:Ablation} as numerical evidence to the significance of the different cost functions, as well as our DGAT module importance. 
While some objectives improve the refinement effect, some, as $\mathcal{L}_{AG}$ or $\mathcal{L}_{l2}$ are indispensable, with substantial degradation to the results without their regularization effect to the denoising process. Inspecting the effect of replacing DGAT with GAT \cite{gat} or GCN \cite{GCN} we witness substantial performance decrease, where the baseline alternatives bring degraded results compared to the initiator mapping.
The ablation was done on FAUST resampled, with the initial correspondence generated by SURFMNet.
\vspace{5mm}

\begin{table}[H]
\begin{tabular}{llc}
&Ablation                          & MGE          \\ \hline
Baseline&SURFMNet & 7.1         \\ \hline

\multirow{5}{*}{Loss}&$\mathcal{L}_L+\mathcal{L}_{l1}$                         & 47.5 \\ 
&$\mathcal{L}_L+\mathcal{L}_{l1}+\mathcal{L}_{AG}$                      & 38.1         \\
&$\mathcal{L}_{l2}+\mathcal{L}_{AG}$                         & 9.9         \\
&$\mathcal{L}_L+\mathcal{L}_{l1}+\mathcal{L}_{l2}$ & 6.7          \\
&$\mathcal{L}_L+\mathcal{L}_{l2}+\mathcal{L}_{AG}$ & 6.2          \\ \cline{1-3}

\multirow{3}{*}{GNN}&DGCNN \cite{dgcnn} & 25.1 \\ 
&GCN \cite{GCN} & 21.9 \\ 
&GAT  \cite{gat} & 14.3 \\\cline{1-3}

Full      & DG2N & \textbf{5.9} \\\hline
\end{tabular}
\caption{Ablation study}
\label{tab:Ablation}
\end{table}



\section{Summary}
\label{sec:summary_future_work}
We presented a novel line of thought for aligning non-rigid domains using a learnable iterative pipeline. Motivated by graph denoising and presenting a dual graph structure built on top of soft correspondences, we rapidly converge into an accurate and free of outliers mapping even under severe non-isometric deformations. We report state-of-the-art results on multiple benchmarks and different scenarios, where other methods suffer poor outcomes or fail altogether.


\newpage
{\small
\bibliographystyle{ieee_fullname}
\bibliography{egbib}
}

\end{document}